%% file: neurips_2026.tex
\documentclass{article}

\usepackage[preprint]{neurips_2026}

\usepackage[utf8]{inputenc}
\usepackage[T1]{fontenc}
\usepackage{hyperref}
\usepackage{url}
\usepackage{booktabs}
\usepackage{amsfonts}
\usepackage{amsmath}
\usepackage{amssymb}
\usepackage{nicefrac}
\usepackage{microtype}
\usepackage{xcolor}
\usepackage{graphicx}
\usepackage{subcaption}
\usepackage{multirow}
\usepackage{array}
\usepackage{colortbl}
\usepackage{algorithm}
\usepackage{algpseudocode}
\usepackage{wrapfig}
\usepackage{tabularx}
\usepackage{array}
\newcommand{\methodname}{TextLDM}

\title{\methodname: Language Modeling with\\Continuous Latent Diffusion}

\author{%
  Jiaxiu Jiang\textsuperscript{1 2}\thanks{Equal contribution.}$\quad$
  Jingjing Ren\textsuperscript{3}\footnotemark[1]$\quad$
  Wenbo Li\textsuperscript{1}\thanks{Corresponding author.}$\quad$
  Bo Wang\textsuperscript{1}$\quad$\\
  \bf
  Haoze Sun\textsuperscript{1}$\quad$
  Yijun Yang\textsuperscript{1}$\quad$
  Jianhui Liu\textsuperscript{1}$\quad$
  Yanbing Zhang\textsuperscript{1}$\quad$
  Shenghe Zheng\textsuperscript{1}$\quad$
  \\
  \bf
  Yuan Zhang\textsuperscript{1}$\quad$
  Haoyang Huang\textsuperscript{1}$\quad$
  Nan Duan\textsuperscript{1}$\quad$
  Wangmeng Zuo\textsuperscript{2}\\
  \textsuperscript{1}Joy Future Academy$\quad$
  \textsuperscript{2}HIT$\quad$
  \textsuperscript{3}HKUST(GZ)$\quad$
}

\begin{document}

\maketitle

\begin{abstract}
Diffusion Transformers (DiT) trained with flow matching in a VAE latent space have unified visual generation across images and videos. A natural next step toward a single architecture for both generation (visual synthesis) and understanding (text generation) is to apply this framework to language modeling. We propose \methodname{}, which transfers the visual latent diffusion recipe to text generation with minimal architectural modification. A Transformer-based VAE maps discrete tokens to continuous latents, enhanced by \emph{Representation Alignment} (REPA) with a frozen pretrained language model to produce representations effective for conditional denoising. A standard DiT then performs flow matching in this latent space, identical in architecture to its visual counterpart. The central challenge we address is obtaining high-quality continuous text representations: we find that reconstruction fidelity alone is insufficient, and that aligning latent features with a pretrained language model via REPA is critical for downstream generation quality. Trained from scratch on OpenWebText2, \methodname{} substantially outperforms prior diffusion language models and matches GPT-2 under the same settings. Our results establish that the visual DiT recipe transfers effectively to language, taking a concrete step toward unified diffusion architectures for multimodal generation and understanding.
\end{abstract}

\section{Introduction}

Central to multimodal modeling is the pursuit of a unified framework capable of seamlessly generating both textual and visual content. On the visual side, Diffusion Transformers (DiT)~\citep{peebles2023scalable} trained with flow matching~\citep{lipman2022flow, liu2022flow} in a VAE latent space~\citep{rombach2022high} have already unified image and video generation~\citep{esser2024scaling, wan2025wan}, establishing a dominant recipe: continuous latent space, DiT backbone, flow matching objective, classifier-free guidance (CFG)~\citep{ho2022classifier}, and carefully designed timestep schedules~\citep{esser2024scaling}. While some autoregressive (AR) methods~\citep{team2024chameleon, wang2024emu3} have attempted to unify understanding and generation within a discrete token-based paradigm, the prevailing excellence of diffusion models in the visual domain still leaves a methodological gap in language modeling. If this same architecture could also perform language modeling effectively, it would provide a concrete foundation for unified multimodal generation and understanding.

\begin{figure}[t]
  \centering
  \includegraphics[width=0.94\linewidth]{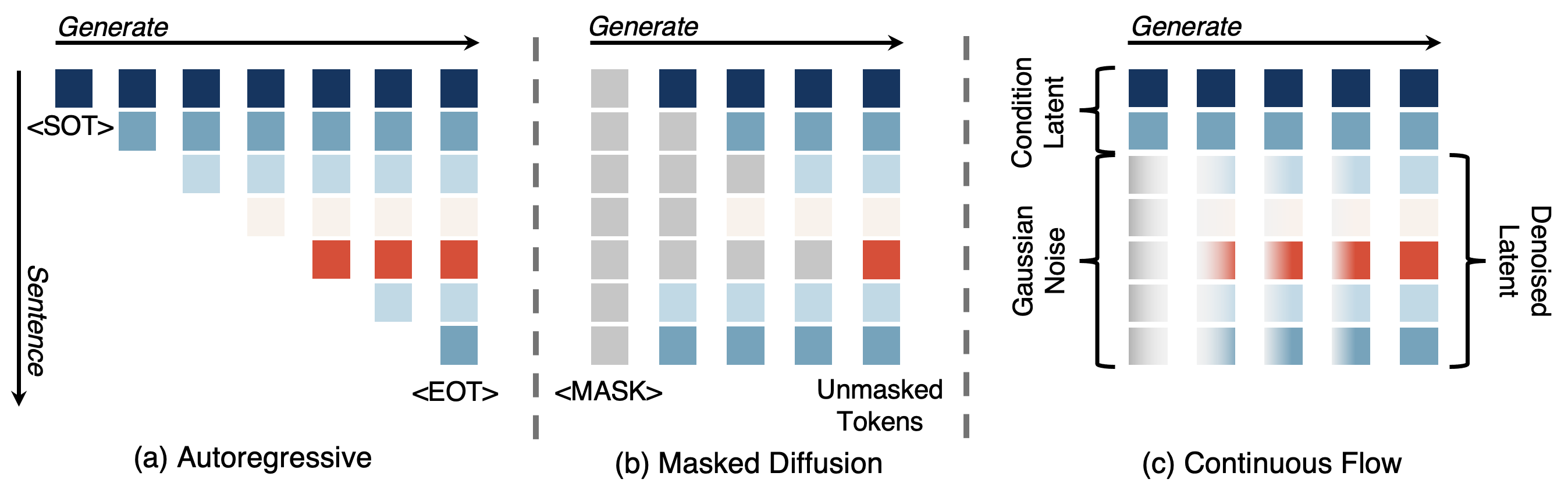}
  \caption{Comparison of language generation paradigms. Each sub-figure illustrates the generation process \emph{from left to right}. \textbf{(a)} Autoregressive (AR) models generate tokens one by one in sequential order. \textbf{(b)} Discrete diffusion (e.g., LLaDA~\citep{nie2025large}) iteratively unmasks tokens, where gray blocks denote \texttt{<mask>} tokens. \textbf{(c)} \methodname{} (ours) performs flow matching in a learned continuous latent space, progressively denoising random noise into valid language representations. The uniform-colored blocks at the bottom represent the condition (context) latents.}
  \vspace{-8pt}
  \label{fig:paradigm_compare}
\end{figure}

\begin{wrapfigure}{r}{0.4\textwidth}
  \centering
  \vspace{-12pt}
  \includegraphics[width=0.38\textwidth]{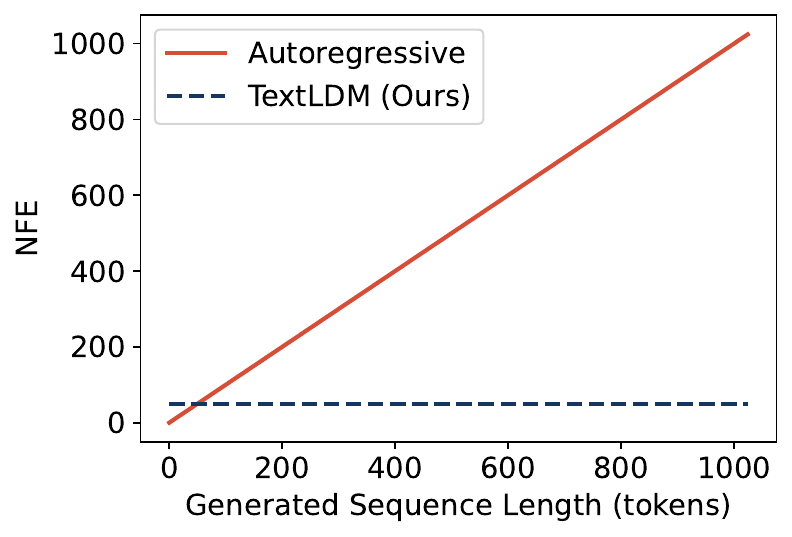}
  \caption{Inference efficiency: the number of function evaluations (NFE) for AR models grows linearly with sequence length, while \methodname{} achieves length-invariant NFE over a broad operating range (e.g., up to 1024 tokens).}
  \label{fig:nfe}
  \vspace{-10pt}
\end{wrapfigure}

As illustrated in Figure~\ref{fig:paradigm_compare}, language generation has traditionally been dominated by the AR paradigm, whereas visual generation has converged toward continuous diffusion modeling. Rather than debating the superiority of one paradigm over the other, this paper explores the feasibility of extending the successful visual diffusion recipe to text generation.
We propose \methodname{}, which instantiates language modeling within the DiT framework with minimal architectural modification. A Transformer-based VAE (\textbf{TextVAE}) maps each discrete token to a continuous latent vector, and a standard DiT (\textbf{TextDiT})---architecturally identical to its visual counterpart---performs flow matching in this latent space. Critically, as shown in Figure~\ref{fig:nfe}, this approach in principle offers a distinct advantage in inference efficiency, providing a more constant-time generation profile.

The central challenge we encounter is not in the diffusion backbone, but in the latent representation. Text is inherently discrete, and a VAE trained solely for token reconstruction can achieve near-perfect accuracy yet produce latents poorly suited for conditional denoising. Our ablations confirm this: configurations with virtually identical reconstruction accuracy may yield substantially different generation quality. The key bottleneck is \emph{representation effectiveness}---whether the continuous latents support the downstream diffusion process---rather than reconstruction fidelity alone.

To address this, we introduce \emph{Representation Alignment} (REPA)~\citep{yu2024representation}, originally proposed for image DiT training, to the text VAE. By aligning the VAE encoder's features with those of a frozen pretrained language model (Qwen3-1.7B~\citep{yang2025qwen3}), REPA shapes the latent geometry to be more amenable to diffusion-based generation, yielding substantial improvements in downstream quality without affecting reconstruction.

We train all components from scratch on OpenWebText2~\citep{pile} and evaluate on text continuation across four benchmarks. \methodname{} substantially outperforms prior continuous and discrete diffusion language models and matches GPT-2 baselines under the same settings. Comprehensive ablations validate that visual diffusion components---logit-normal scheduling and CFG---can seamlessly transfer to the language modeling.

Our contributions are:
\begin{itemize}
    \item We propose \methodname{}, which transfers the visual latent diffusion recipe (VAE + DiT + flow matching + CFG) to language modeling with minimal modification, taking a concrete step toward unified diffusion architectures for multimodal generation and understanding. The entire system is trained from scratch without pretrained encoders or decoders.
    \item We identify representation effectiveness as the key bottleneck for latent text diffusion, and introduce REPA-enhanced TextVAE to produce continuous representations suited for conditional denoising, substantially improving generation quality without affecting reconstruction.
    \item Extensive experiments demonstrate that \methodname{} achieves state-of-the-art results among diffusion language models on text continuation benchmarks and matches autoregressive baselines under identical settings. Ablations validate the effectiveness of each transferred visual diffusion component.
\end{itemize}

\section{Related Work}

\paragraph{Diffusion Models for Visual Generation.}
Diffusion models~\citep{ho2020denoising, song2020denoising} have been unified with flow matching~\citep{liu2022flow, lipman2022flow} and extended to latent spaces~\citep{rombach2022high}. Diffusion Transformers (DiT)~\citep{peebles2023scalable} enabled scalable architectures, and the recipe of flow matching + VAE + DiT has become the standard for visual generation~\citep{esser2024scaling, chen2024pixart}. 

\paragraph{Diffusion Language Models.}
Diffusion language models (see Figure~\ref{fig:paradigm_compare}) can be categorized into continuous and discrete approaches. Continuous methods~\citep{li2022diffusion, lin2023text, wu2023ar, han2023ssd, dieleman2022continuous, gulrajani2023likelihood} apply diffusion in embedding or simplex spaces. LD4LG~\citep{lovelace2023latent} and COSMOS~\citep{meshchaninovcosmos} use latent diffusion with pretrained encoders or compressed latent spaces. Discrete methods~\citep{austin2021structured, sahoo2024simple, nie2025large} define diffusion over tokens directly. Block Diffusion~\citep{arriola2025block} denoises token blocks, and CALM~\citep{shao2025continuous} augments AR with a diffusion head for chunk generation. Our method differs by performing flow matching in a learned continuous latent space with a standard DiT, requiring no pretrained encoder/decoder. Unlike chunk-based methods, we generate an entire passage in a single diffusion pass.

\paragraph{Variational Autoencoders for Text.}
Prior text VAEs~\citep{kingma2013auto, li2020optimus, liu2019mu} typically rely on pretrained components or autoregressive decoders. Our TextVAE is trained from scratch with a non-autoregressive decoder and enhanced by REPA~\citep{yu2024representation}, which was originally proposed to align DiT representations with pretrained vision encoders for image generation. We adapt REPA to align the VAE encoder with a frozen language model, ensuring the latent space is highly structured and semantically rich, which significantly enhances representation quality.

\section{Method}
\label{sec:method}

\begin{figure}[t]
  \centering
  \includegraphics[width=0.93\linewidth]{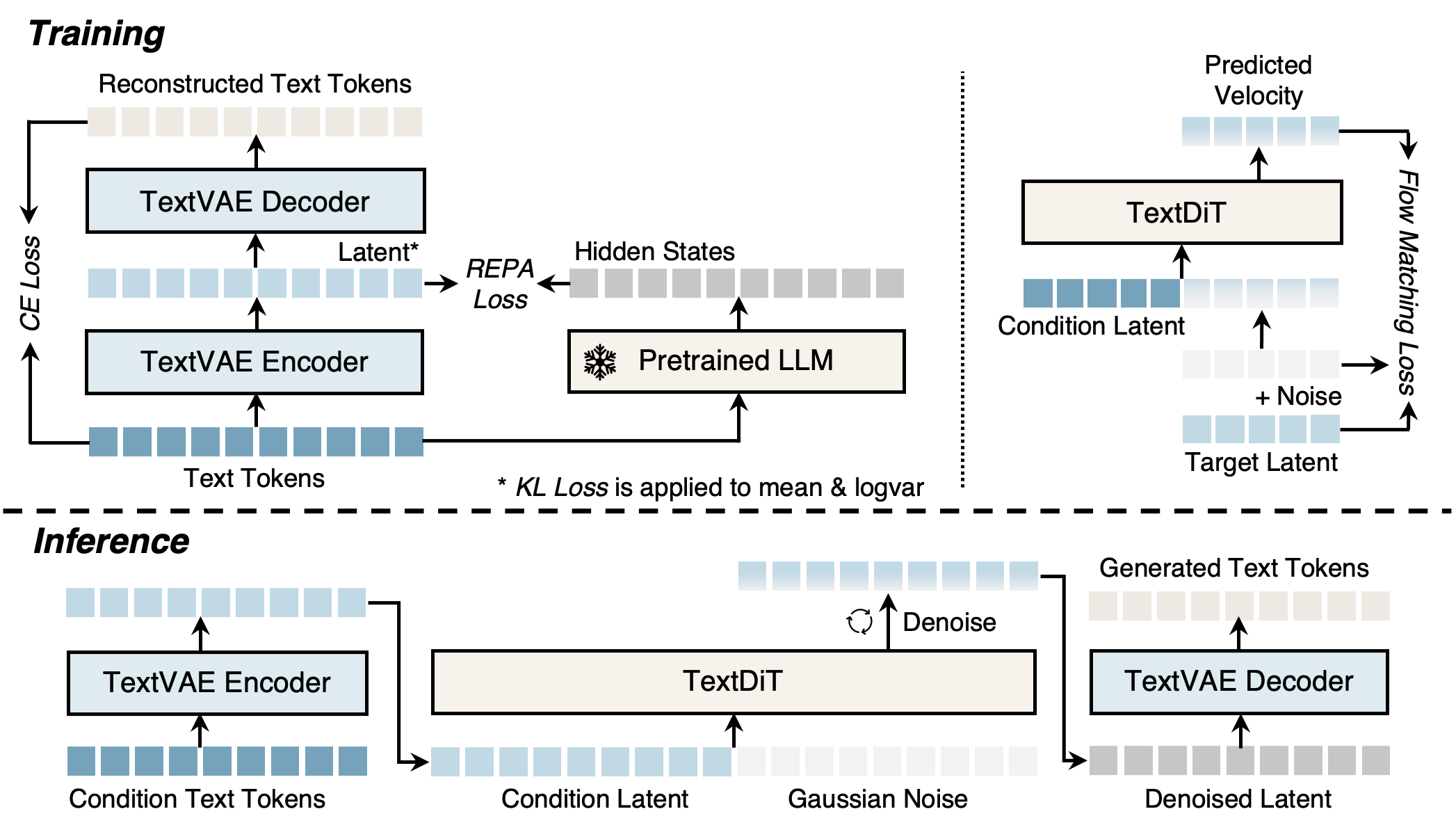}
  \caption{Overview of \methodname{}. \textbf{(a) TextVAE}: A Transformer encoder maps discrete tokens to continuous latents, regularized by KL divergence and enhanced by REPA alignment with a frozen Qwen3-1.7B. The decoder reconstructs tokens from latents via cross-entropy loss. \textbf{(b) TextDiT}: A Diffusion Transformer is trained with Flow Matching. Clean context latents and noisy target latents are concatenated as input; the model predicts the velocity field to denoise the target segment conditioned on the context. For unconditional generation, the condition latent is omitted.}
  \label{fig:method_overview}
\end{figure}

We present \methodname{}, a two-stage framework for language modeling through continuous latent diffusion. As illustrated in Figure~\ref{fig:method_overview}, the framework consists of (a) a TextVAE that compresses discrete text tokens into continuous latent representations, and (b) a Diffusion Transformer trained with Flow Matching to model generative dynamics in the latent space.

\subsection{TextVAE: Continuous Latent Representations for Text}

\paragraph{Architecture.}
Let $\mathbf{x} = (x_1, x_2, \dots, x_N)$ denote a sequence of $N$ discrete tokens obtained by a standard tokenizer (we use the Qwen3 tokenizer~\citep{yang2025qwen3}), where $x_i \in \mathcal{V}$ and $\mathcal{V}$ is the vocabulary. Unlike prior latent diffusion approaches for text that compress the token sequence into a shorter latent sequence~\citep{lovelace2023latent, meshchaninovcosmos}, our TextVAE maintains a \emph{one-to-one mapping}: each token $x_i$ corresponds to exactly one latent vector $\mathbf{z}_i \in \mathbb{R}^d$, where $d$ is the latent channel dimension.

The encoder $E_\phi$ is a Transformer that processes the input tokens and produces parameters of a diagonal Gaussian posterior for each position:
\begin{equation}
q_\phi(\mathbf{z}_i \mid \mathbf{x}) = \mathcal{N}(\boldsymbol{\mu}_i, \boldsymbol{\sigma}_i^2), \quad i = 1, \dots, N
\end{equation}
where $\boldsymbol{\mu}_i, \boldsymbol{\sigma}_i$ are predicted by the encoder. Latent vectors are sampled via the reparameterization trick: $\mathbf{z}_i = \boldsymbol{\mu}_i + \boldsymbol{\sigma}_i \odot \boldsymbol{\epsilon}$, $\boldsymbol{\epsilon} \sim \mathcal{N}(\mathbf{0}, \mathbf{I})$.

The decoder $D_\psi$ is also a Transformer that takes the latent sequence $\mathbf{z} = (\mathbf{z}_1, \dots, \mathbf{z}_N)$ as input and predicts a probability distribution over the vocabulary for each position, reconstructing the original tokens in parallel (non-autoregressively). During VAE training, input sequences are randomly truncated so that the model learns to reconstruct varying portions and lengths. When training the downstream DiT, the context and target segments are encoded \emph{separately} by the VAE encoder, rather than encoding the full sequence and splitting afterward, to prevent information leakage from target tokens into the context latents.

\paragraph{Representation Alignment (REPA).}
To enrich the VAE latent space with the semantic knowledge captured by pretrained language models, we introduce Representation Alignment~\citep{yu2024representation} to text VAE training. We leverage a frozen pretrained language model---specifically Qwen3-1.7B~\citep{yang2025qwen3}---as a representation target, aligning the VAE encoder's intermediate representations with the LLM's hidden states via a cosine similarity loss:
\begin{equation}
\mathcal{L}_{\text{REPA}} = -\frac{1}{N} \sum_{i=1}^{N} \text{cos}(\mathbf{h}_i^{\text{enc}}, \text{sg}(\mathbf{h}_i^{\text{LLM}}))
\label{eq:repa}
\end{equation}
where $\mathbf{h}_i^{\text{enc}}$ denotes the encoder's intermediate representation at position $i$, $\mathbf{h}_i^{\text{LLM}}$ denotes the corresponding representation from the frozen language model, and $\text{sg}(\cdot)$ denotes the stop-gradient operation. A linear projection layer is applied to match dimensions when necessary.

In our experiments, we align the encoder's output with representations from the 3rd-to-last layer of Qwen3-1.7B, which we find works better than the last layer (see ablation in Section~\ref{sec:ablation}).

\paragraph{Training Objective.}
The TextVAE is trained with a composite loss:
\begin{equation}
\mathcal{L}_{\text{VAE}} = \mathcal{L}_{\text{CE}}(\mathbf{x}, \hat{\mathbf{x}}) + \beta \cdot D_{\text{KL}}\big(q_\phi(\mathbf{z} \mid \mathbf{x}) \,\|\, \mathcal{N}(\mathbf{0}, \mathbf{I})\big) + \lambda \cdot \mathcal{L}_{\text{REPA}}
\end{equation}
where $\mathcal{L}_{\text{CE}}$ is the cross-entropy reconstruction loss, $D_{\text{KL}}$ regularizes the latent posterior toward a standard Gaussian prior, and $\mathcal{L}_{\text{REPA}}$ enforces representation alignment. We set $\beta = 10^{-3}$ and $\lambda = 1$. After training, the encoder produces a smooth, semantically rich latent space suitable for diffusion-based generation.

\subsection{Latent Diffusion via Flow Matching}

After training the TextVAE, we freeze the encoder and train a Diffusion Transformer (DiT) in the learned latent space using Flow Matching~\citep{lipman2022flow, liu2022flow}.

\paragraph{Conditional Formulation.}
To model language generation as a conditional process, we divide the latent sequence into two parts:
\begin{itemize}
    \item \textbf{Context} $\mathbf{z}_c = (\mathbf{z}_1, \dots, \mathbf{z}_M)$: latent representations of the preceding text (the ``prompt'').
    \item \textbf{Target} $\mathbf{z}_{\text{tgt}} = (\mathbf{z}_{M+1}, \dots, \mathbf{z}_N)$: latent representations of the text to be generated.
\end{itemize}
The model learns the conditional distribution $p(\mathbf{z}_{\text{tgt}} \mid \mathbf{z}_c)$, generating the entire target segment simultaneously via the diffusion process. To also enable unconditional generation of full passages, we set $\mathbf{z}_{\text{tgt}} = (\mathbf{z}_1, \dots, \mathbf{z}_N)$ with no context with probability $10\%$ during training.

\paragraph{Flow Matching Objective.}
We construct the noisy intermediate state by linearly interpolating between Gaussian noise and the target latent:
\begin{equation}
\mathbf{z}_t = (1 - t) \mathbf{z}_0 + t \, \mathbf{z}_{\text{tgt}}, \quad t \in [0, 1], \quad \mathbf{z}_0 \sim \mathcal{N}(\mathbf{0}, \mathbf{I})
\end{equation}
The DiT $v_\theta$ takes as input the concatenation of clean context latents $\mathbf{z}_c$ and noisy target latents $\mathbf{z}_t$, along with the timestep $t$, and predicts the velocity field. The model is optimized with the Conditional Flow Matching (CFM) loss:
\begin{equation}
\mathcal{L}_{\text{FM}} = \mathbb{E}_{t, \mathbf{z}_0, \mathbf{z}_{\text{tgt}}} \left[ \left\| v_\theta(\mathbf{z}_t, t, \mathbf{z}_c) - (\mathbf{z}_{\text{tgt}} - \mathbf{z}_0) \right\|^2 \right]
\end{equation}
where the timestep $t$ is sampled from a logit-normal distribution, following the finding from Stable Diffusion 3~\citep{esser2024scaling} that this schedule provides better training signal distribution than a uniform schedule. Following CDCD~\citep{dieleman2022continuous}, we use the same timestep scheduler for both training and inference.

\paragraph{Classifier-Free Guidance.}
We apply classifier-free guidance (CFG)~\citep{ho2022classifier} to improve generation quality. During training, context latents $\mathbf{z}_c$ are randomly replaced with zero vectors with probability $p_{\text{uncond}}=0.1$. At inference, the guided velocity is:
\begin{equation}
\tilde{v}_\theta = v_\theta(\mathbf{z}_t, t, \varnothing) + w \cdot \big(v_\theta(\mathbf{z}_t, t, \mathbf{z}_c) - v_\theta(\mathbf{z}_t, t, \varnothing)\big)
\end{equation}
where $w$ is the guidance scale and $\varnothing$ denotes the null condition.

\subsection{Inference}

\begin{algorithm}[h]
\caption{Inference of \methodname{}}
\label{alg:inference}
\begin{algorithmic}[1]
\Require Text prompt $\mathbf{x}_{\text{prompt}}$, TextVAE encoder $E_\phi$ and decoder $D_\psi$, TextDiT $v_\theta$, CFG scale $w$, timestep schedule $1 = t_K > t_{K-1} > \cdots > t_0 = 0$
\State $\mathbf{z}_c \gets E_\phi(\mathbf{x}_{\text{prompt}})$ \Comment{Encode prompt into context latents}
\State $\mathbf{z}_{t_K} \sim \mathcal{N}(\mathbf{0}, \mathbf{I})$ \Comment{Sample noise for target positions}
\For{$k = K, K-1, \dots, 1$} \Comment{Euler ODE solver}
    \State $\Delta t \gets t_k - t_{k-1}$
    \State $\tilde{v} \gets v_\theta(\mathbf{z}_{t_k}, t_k, \varnothing) + w \cdot \big(v_\theta(\mathbf{z}_{t_k}, t_k, \mathbf{z}_c) - v_\theta(\mathbf{z}_{t_k}, t_k, \varnothing)\big)$ \Comment{CFG}
    \State $\mathbf{z}_{t_{k-1}} \gets \mathbf{z}_{t_k} - \Delta t \cdot \tilde{v}$
\EndFor
\State $\hat{\mathbf{x}}_{\text{tgt}} \gets D_\psi(\mathbf{z}_{t_0})$ \Comment{Decode latents to tokens}
\State \Return $\hat{\mathbf{x}}_{\text{tgt}}$
\end{algorithmic}
\end{algorithm}

The entire target segment is generated in parallel, avoiding token-level autoregressive decoding. For unconditional generation, the context encoding step is skipped and $\varnothing$ is used throughout.

\begin{table*}[t]
  \caption{Text continuation results across four benchmarks. All models are trained on OpenWebText2 with max sequence length 1024. Our \methodname{} uses the default configuration: VAE 350M with ch64 and REPA (Qwen3-1.7B, 3rd-to-last layer), logit-normal 1.5 scheduler, CFG=7, 50-step inference. Best results per column are \textbf{bolded}.}
  \label{tab:main_results}
  \centering
  \renewcommand{\arraystretch}{1.1}
  \setlength{\tabcolsep}{3pt}
  \resizebox{\textwidth}{!}{
  \begin{tabular}{l|ccccc|ccccc|ccccc|ccccc}
    \toprule
    & \multicolumn{5}{c|}{WikiSource} & \multicolumn{5}{c|}{Wikipedia} & \multicolumn{5}{c|}{TinyStories} & \multicolumn{5}{c}{One Billion Words} \\
    \cmidrule(lr){2-6} \cmidrule(lr){7-11} \cmidrule(lr){12-16} \cmidrule(lr){17-21}
    Model & R-1 & R-2 & R-L & BS & MAU & R-1 & R-2 & R-L & BS & MAU & R-1 & R-2 & R-L & BS & MAU & R-1 & R-2 & R-L & BS & MAU \\
    \midrule
    \multicolumn{21}{l}{\textit{Autoregressive Models}} \\
    GPT-2 (137M) & 31.1 & 7.0 & 18.2 & 81.6 & 35.3 & 23.3 & 4.7 & 15.1 & 81.6 & 7.89 & 31.8 & 6.1 & 18.9 & 85.5 & 1.04 & 13.4 & 1.6 & 12.3 & 83.9 & 0.45 \\
    GPT-2-medium (355M) & 34.0 & 8.3 & 19.0 & 82.4 & 39.2 & 25.0 & 5.3 & 15.7 & 81.8 & 8.16 & 33.6 & 6.8 & 19.9 & 86.1 & 1.47 & 14.8 & 2.4 & 13.8 & 84.2 & 0.49 \\
    GPT-2-large (774M) & 33.7 & 8.4 & 19.5 & 82.2 & 38.3 & 25.1 & 5.7 & 16.1 & 82.2 & 8.00 & 34.7 & 7.5 & 20.8 & \textbf{86.3} & 1.47 & 15.8 & 2.9 & 14.6 & 84.3 & 0.53 \\
    \midrule
    \multicolumn{21}{l}{\textit{Discrete Diffusion Language Models}} \\
    BlockDiff. (170M) bs=16 & 30.9 & 5.6 & 16.0 & 81.4 & 39.6 & 23.9 & 4.2 & 13.8 & 81.6 & 7.23 & 31.8 & 5.3 & 18.0 & 85.1 & 1.20 & 10.9 & 0.6 & 9.4 & 83.1 & 0.50 \\
    BlockDiff. (170M) bs=8 & 30.9 & 5.6 & 16.0 & 81.5 & \textbf{41.0} & 22.9 & 4.1 & 13.4 & 81.3 & 7.70 & 31.0 & 5.0 & 17.7 & 84.9 & 1.00 & 9.0 & 0.7 & 8.1 & 82.8 & 0.55 \\
    BlockDiff. (170M) bs=4 & 29.3 & 5.2 & 15.3 & 81.1 & 29.8 & 21.9 & 3.8 & 13.1 & 80.9 & 7.30 & 28.5 & 4.4 & 16.6 & 84.4 & 1.10 & 9.5 & 0.8 & 8.7 & 83.0 & 0.59 \\
    \midrule
    \multicolumn{21}{l}{\textit{Continuous Diffusion Language Models}} \\
    SSD-LM (355M) & 15.3 & 2.0 & 10.4 & 79.2 & 7.66 & 15.1 & 1.8 & 10.4 & 79.3 & 2.82 & 28.0 & 3.7 & 16.6 & 83.7 & 0.78 & 10.0 & 0.7 & 8.8 & 82.8 & 0.65 \\
    Ours--\methodname{} (114M) & 33.0 & 6.6 & 16.6 & 80.3 & 21.6 & 27.5 & 5.9 & 15.9 & 81.0 & 8.9 & 36.7 & 7.8 & 20.7 & 84.8 & 1.00 & 10.3 & 0.73 & 9.4 & 83.1 & 0.77 \\
    Ours--\methodname{} (328M) & 33.1 & 6.8 & 16.9 & 80.7 & 27.6 & 27.6 & 6.2 & 16.2 & 81.3 & \textbf{10.5} & 37.1 & 8.3 & 21.1 & 85.2 & 1.13 & 10.8 & 0.88 & 9.8 & 83.4 & 0.79 \\
    Ours--\methodname{} (768M) & \textbf{37.5} & \textbf{16.5} & \textbf{25.7} & \textbf{84.3} & 32.7 & \textbf{38.9} & \textbf{8.1} & \textbf{17.6} & \textbf{82.7} & 10.1 & \textbf{39.7} & \textbf{10.4} & \textbf{23.4} & 85.8 & \textbf{1.51} & \textbf{21.4} & \textbf{3.6} & \textbf{17.4} & \textbf{85.0} & \textbf{0.80} \\
    \bottomrule
  \end{tabular}
  }
\end{table*}

\section{Experiments}
\label{sec:experiments}

\subsection{Experimental Setup}

\paragraph{Training Data.}
All models are trained on OpenWebText2~\citep{pile}, with a maximum sequence length of 1024 tokens.

\paragraph{Model Configurations.}
For the TextVAE, we experiment with three model sizes (350M, 502M, 690M parameters), latent channel dimensions $d \in \{64, 128, 192\}$, and REPA alignment using the 1st- or 3rd-to-last layer of Qwen3-1.7B~\citep{yang2025qwen3}. Note that 223M of each VAE's parameters are token embeddings and LM head weights. The Transformer encoder and decoder blocks account for the remaining parameters. The VAE is trained for 200K steps. For the latent DiT, we evaluate four model sizes: 114M, 328M, and 768M parameters. The DiTs in ablation study are trained for 1M steps with the logit-normal timestep schedule (std=1.5) unless otherwise noted. The DiTs in Table~\ref{tab:main_results} are trained for 2M steps.

\paragraph{Evaluation.}
We evaluate on the text continuation task across four benchmarks that span a range of difficulty and domain overlap with the training data. One Billion Words~\citep{chelba2014one} consists of short sentences averaging only a few dozen tokens, providing a relatively easy in-domain test. TinyStories~\citep{eldan2023tinystories} contains slightly longer samples but is restricted to simple children's stories with limited topical diversity. Wikipedia\footnote{\url{https://huggingface.co/datasets/wikimedia/wikipedia}} and WikiSource\footnote{\url{https://huggingface.co/datasets/wikimedia/wikisource}} contain substantially longer documents with highly diverse content that is out-of-distribution with respect to OpenWebText2, thus testing generalization ability.

For each benchmark, we randomly sample 1K test examples (truncated to 1024 tokens if longer). Each sample is split into a condition prefix and a ground-truth target at a split point uniformly drawn between 40\% and 60\% of the sample length, ensuring diverse condition and target lengths. The condition prefix is fed to the model, and the generated continuation is compared against the ground-truth target. We report ROUGE-1, ROUGE-2, ROUGE-L~\citep{lin2004rouge}, BERTScore~\citep{zhang2020bertscore}, and MAUVE~\citep{pillutla2021mauve}. At inference, we use 50-step Euler sampling with CFG scale $w=7$ unless otherwise noted.

\paragraph{Baselines.}
We compare against: (1) \textbf{AR models}: Pretrained GPT-2 (137M, 355M, 774M)~\citep{radford2019language}; (2) \textbf{Continuous diffusion LMs}: SSD-LM (355M)~\citep{han2023ssd} trained on OpenWebText~\citep{Gokaslan2019OpenWeb}; (3) \textbf{Discrete diffusion LMs}: Block Diffusion (170M)~\citep{arriola2025block} with block sizes 4, 8, and 16 trained on OpenWebText.
We note that several recent diffusion LMs are excluded from our comparison for fairness. PLAID~\citep{gulrajani2023likelihood} and COSMOS~\citep{meshchaninovcosmos} only release checkpoints for unconditional generation, which does not align with our text continuation evaluation protocol. CALM~\citep{shao2025continuous} and LLaDA~\citep{nie2025large} are trained on substantially larger corpora than OpenWebText2, making direct comparison inequitable.

\begin{figure*}[t]
  \centering
  \includegraphics[width=0.95\textwidth]{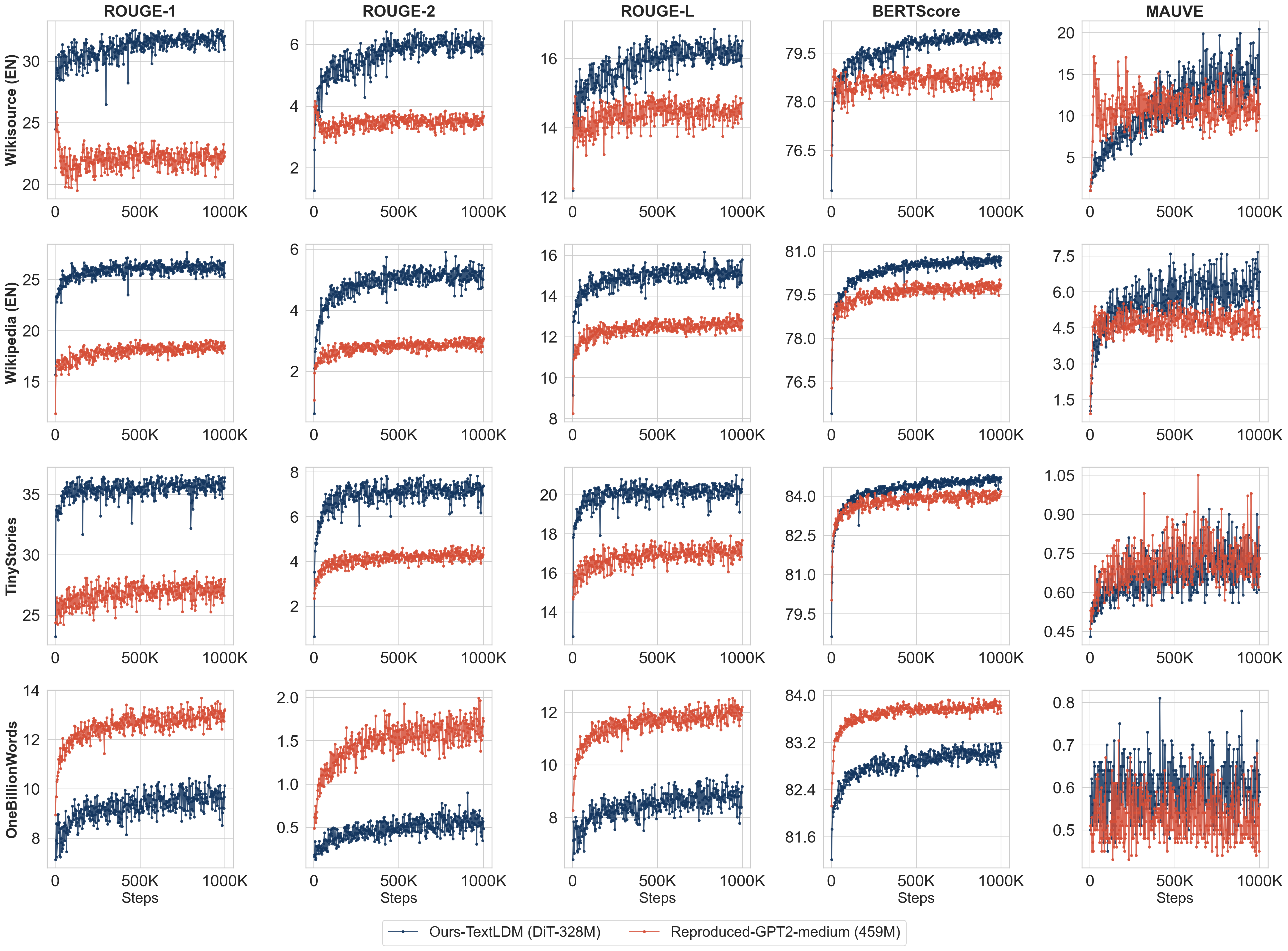}
  \caption{Training dynamics comparison between \methodname{} (DiT-328M, \textcolor[RGB]{20,54,95}{blue}) and our reproduced GPT-2-medium (459M, \textcolor[RGB]{214,79,56}{red}). The reproduced GPT-2-medium has more parameters than the original 355M due to the larger vocabulary size of the Qwen3 tokenizer. Both models are trained from scratch on OpenWebText2 with the same Qwen3 tokenizer and evaluated at identical checkpoint intervals. Each dot represents a checkpoint evaluation.
  }
  \label{fig:training_curve}
\end{figure*}

\subsection{Main Results}

Table~\ref{tab:main_results} presents the main results. Several observations emerge:

\textbf{\methodname{} significantly outperforms prior diffusion language models.} Compared to SSD-LM and Block Diffusion, \methodname{} achieves substantial improvements across all ROUGE, BERTScore, and MAUVE metrics on all four benchmarks, even at comparable model size. Notably, our 768M model achieves the best results on the majority of metrics, surpassing all baselines including GPT-2 models.

\textbf{\methodname{} achieves superior or comparable performance to AR baselines.} On TinyStories and One Billion Words, \methodname{} matches or exceeds GPT-2 models of similar or even larger size on ROUGE metrics. On the more challenging out-of-distribution benchmarks (Wikipedia and WikiSource), our model remains competitive with GPT-2, with the 768M variant outperforming all GPT-2 models. The remaining gap on some benchmarks is primarily on BERTScore, where AR models retain an advantage.

\textbf{Consistent scaling behavior.} \methodname{} shows clear improvements when scaling from 114M to 768M across all metrics. On MAUVE, which measures distributional similarity to human text, the 768M model achieves 32.7 on WikiSource (vs.\ 21.6 for 114M) and 1.51 on TinyStories (vs.\ 1.00 for 114M). ROUGE-1 likewise improves consistently, e.g., from 33.0 to 37.5 on WikiSource and from 10.3 to 21.4 on One Billion Words. The 768M variant also shows a notable jump on Wikipedia (R-1: 27.5$\to$38.9), suggesting that larger models better capture long-range coherence required for encyclopedic text. These trends indicate that the continuous latent diffusion paradigm benefits from increased model capacity in a manner similar to autoregressive language models.

\textbf{Comparable training efficiency to AR models.} Figure~\ref{fig:training_curve} compares the training dynamics of \methodname{} (DiT-328M) and GPT-2-medium (459M) under identical settings: both are trained from scratch on OpenWebText2 with the same Qwen3 tokenizer, evaluated at the same checkpoint intervals. On WikiSource, Wikipedia, and TinyStories, \methodname{} matches or exceeds GPT-2-medium on ROUGE and MAUVE within a comparable number of training steps. On One Billion Words, however, our model lags slightly behind. We hypothesize this is because One Billion Words consists of very short samples, and our training procedure uniformly samples sequence lengths, resulting in relatively few short-sample training instances. In contrast, autoregressive models effectively observe all prefix lengths for every sample at each training step, giving them a natural advantage on short-text benchmarks. Increasing the sampling probability for short sequences could potentially close this gap. Conversely, the strong performance on longer-document benchmarks suggests that diffusion models may hold an advantage for long-range text modeling. Overall, these results demonstrate that continuous latent diffusion can achieve training efficiency on par with autoregressive models---a significant improvement over prior continuous diffusion language models, which typically require substantially more compute to reach comparable quality.

\subsection{Ablation Study}
\label{sec:ablation}

We conduct comprehensive ablations to validate key design choices. Unless otherwise noted, ablations use the 328M DiT with VAE 350M (ch128, REPA Qwen3-1.7B 3rd-to-last layer, logit-normal 1.5). Results are summarized in Table~\ref{tab:ablation}.

\begin{table*}[t]
  \caption{Ablation studies on downstream DiT generation performance under different design choices. Unless otherwise noted, DiTs are trained for 1M steps with CFG=7 and 50 inference steps. Group (e) reports results at 2M steps as we found 1M steps insufficient for convergence at these scales. Default configuration: VAE 350M, ch128, REPA (Qwen3-1.7B, 3rd-to-last layer), DiT 328M, logit-normal 1.5. Best results per group are \textbf{bolded}.}
  \label{tab:ablation}
  \centering
    \renewcommand{\arraystretch}{1.1}
  \setlength{\tabcolsep}{3pt}
  \resizebox{\textwidth}{!}{
  \begin{tabular}{l|ccccc|ccccc|ccccc|ccccc}
    \toprule
    & \multicolumn{5}{c|}{WikiSource} & \multicolumn{5}{c|}{Wikipedia} & \multicolumn{5}{c|}{TinyStories} & \multicolumn{5}{c}{One Billion Words} \\
    \cmidrule(lr){2-6} \cmidrule(lr){7-11} \cmidrule(lr){12-16} \cmidrule(lr){17-21}
    Setting & R-1 & R-2 & R-L & BS & MAU & R-1 & R-2 & R-L & BS & MAU & R-1 & R-2 & R-L & BS & MAU & R-1 & R-2 & R-L & BS & MAU \\
    \midrule
    \multicolumn{21}{l}{\textit{(a) Effect of REPA}} \\
    w/ REPA & \textbf{32.6} & \textbf{6.5} & \textbf{16.9} & \textbf{80.2} & \textbf{20.4} & \textbf{27.7} & \textbf{5.9} & \textbf{16.2} & \textbf{81.0} & \textbf{7.7} & \textbf{36.6} & \textbf{7.9} & \textbf{21.0} & \textbf{84.8} & \textbf{0.92} & \textbf{10.5} & \textbf{0.90} & \textbf{9.6} & \textbf{83.2} & \textbf{0.81} \\
    w/o REPA & 27.8 & 4.4 & 14.6 & 77.3 & 2.5 & 22.5 & 3.2 & 13.2 & 77.7 & 4.4 & 35.1 & 5.7 & 19.5 & 81.3 & 0.58 & 9.1 & 0.48 & 8.1 & 82.1 & 0.73 \\
    \midrule
    \multicolumn{21}{l}{\textit{(b) VAE Model Size}} \\
    350M & 32.6 & 6.5 & \textbf{16.9} & 80.2 & 20.4 & \textbf{27.7} & \textbf{5.9} & \textbf{16.2} & \textbf{81.0} & 7.7 & \textbf{36.6} & \textbf{7.9} & \textbf{21.0} & 84.8 & 0.92 & \textbf{10.5} & \textbf{0.90} & \textbf{9.6} & \textbf{83.2} & \textbf{0.81} \\
    502M & 31.9 & 6.3 & 16.3 & 80.2 & 21.7 & 26.7 & 5.5 & 15.5 & 80.9 & 7.4 & 36.4 & 7.7 & 20.8 & \textbf{84.9} & 0.82 & 10.4 & 0.71 & \textbf{9.6} & 83.1 & 0.70 \\
    690M & \textbf{33.2} & \textbf{6.6} & 16.7 & \textbf{80.3} & \textbf{21.9} & 26.5 & 5.5 & 15.3 & 80.8 & \textbf{8.4} & 36.1 & 7.6 & 20.5 & 84.8 & \textbf{1.03} & 10.0 & 0.77 & 9.1 & \textbf{83.2} & 0.70 \\
    \midrule
    \multicolumn{21}{l}{\textit{(c) Latent Channel Dimension}} \\
    32 & 32.6 & 6.7 & 16.5 & \textbf{80.6} & 26.8 & 27.1 & 5.8 & 15.6 & 81.0 & 8.7 & 36.4 & 7.8 & 20.5 & 85.0 & \textbf{1.05} & 9.9 & 0.69 & 9.1 & 83.1 & \textbf{0.81} \\
    64 & \textbf{34.3} & \textbf{7.0} & 17.1 & \textbf{80.6} & \textbf{27.3} & 27.4 & \textbf{5.9} & 16.1 & \textbf{81.1} & \textbf{12.5} & \textbf{37.1} & \textbf{8.2} & \textbf{21.1} & \textbf{85.1} & 0.95 & \textbf{12.7} & 0.93 & \textbf{11.9} & \textbf{83.3} & 0.73 \\
    128 & 32.6 & 6.5 & 16.9 & 80.2 & 20.4 & \textbf{27.7} & \textbf{5.9} & \textbf{16.2} & 81.0 & 7.7 & 36.6 & 7.9 & 21.0 & 84.8 & 0.92 & 10.5 & 0.90 & 9.6 & 83.2 & \textbf{0.81} \\
    192 & 33.8 & 6.6 & \textbf{17.4} & 80.2 & 17.6 & 27.0 & 5.5 & 15.6 & 80.8 & 7.5 & 36.3 & 7.4 & 20.6 & 84.7 & 0.85 & 12.1 & \textbf{1.21} & 11.2 & 83.2 & 0.68 \\
    \midrule
    \multicolumn{21}{l}{\textit{(d) REPA Layer Selection}} \\
    Layer -1 & \textbf{33.1} & 6.3 & 16.8 & 80.1 & 16.5 & 27.1 & 5.5 & 15.7 & 80.8 & 7.5 & \textbf{37.2} & \textbf{8.1} & \textbf{21.2} & 84.8 & \textbf{0.98} & 10.0 & 0.65 & 9.1 & 83.0 & 0.70 \\
    Layer -3 & 32.6 & \textbf{6.5} & \textbf{16.9} & \textbf{80.2} & \textbf{20.4} & \textbf{27.7} & \textbf{5.9} & \textbf{16.2} & \textbf{81.0} & \textbf{7.7} & 36.6 & 7.9 & 21.0 & \textbf{84.8} & 0.92 & \textbf{10.5} & \textbf{0.90} & \textbf{9.6} & \textbf{83.2} & \textbf{0.81} \\
    \midrule
    \multicolumn{21}{l}{\textit{(e) DiT Model Size}} \\
    114M & 33.0 & 6.6 & 16.6 & 80.3 & 21.6 & 27.5 & 5.9 & 15.9 & 81.0 & 8.9 & 36.7 & 7.8 & 20.7 & 84.8 & 1.00 & 10.3 & 0.73 & 9.4 & 83.1 & 0.77 \\
    328M & 33.1 & 6.8 & 16.9 & 80.7 & 27.6 & 27.6 & 6.2 & 16.2 & 81.3 & \textbf{10.5} & 37.1 & 8.3 & 21.1 & 85.2 & 1.13 & 10.8 & 0.88 & 9.8 & 83.4 & 0.79 \\
    768M & \textbf{37.5} & \textbf{16.5} & \textbf{25.7} & \textbf{84.3} & \textbf{32.7} & \textbf{38.9} & \textbf{8.1} & \textbf{17.6} & \textbf{82.7} & 10.1 & \textbf{39.7} & \textbf{10.4} & \textbf{23.4} & \textbf{85.8} & \textbf{1.51} & \textbf{21.4} & \textbf{3.6} & \textbf{17.4} & \textbf{85.0} & \textbf{0.80} \\
    \midrule
    \multicolumn{21}{l}{\textit{(f) Timestep Schedule}} \\
    Uniform & 32.2 & 6.4 & 16.5 & \textbf{80.3} & \textbf{23.2} & 26.8 & 5.6 & 15.4 & 80.9 & 7.1 & 36.3 & 7.7 & 20.7 & \textbf{84.9} & 0.85 & 10.0 & 0.71 & 9.1 & 83.1 & 0.72 \\
    Logit-normal 1.2 & 29.5 & 4.7 & 15.3 & 79.3 & 12.5 & 25.9 & 4.6 & 14.9 & 80.4 & 7.1 & 35.5 & 7.1 & 20.6 & 84.6 & 0.88 & 10.3 & 0.75 & 9.4 & \textbf{83.2} & 0.74 \\
    Logit-normal 1.5 & \textbf{32.6} & \textbf{6.5} & \textbf{16.9} & 80.2 & 20.4 & \textbf{27.7} & \textbf{5.9} & \textbf{16.2} & \textbf{81.0} & \textbf{7.7} & \textbf{36.6} & \textbf{7.9} & \textbf{21.0} & 84.8 & \textbf{0.92} & \textbf{10.5} & \textbf{0.90} & \textbf{9.6} & \textbf{83.2} & \textbf{0.81} \\
    \bottomrule
  \end{tabular}
  }
\end{table*}

\paragraph{Effect of REPA.}
REPA provides substantial improvements across all metrics and all four datasets (group a). The gains are particularly pronounced on the out-of-distribution benchmarks (Wikipedia and WikiSource), demonstrating that aligning the VAE encoder with a pretrained language model significantly enriches the latent space semantics.

\paragraph{VAE Model Size.}
Increasing VAE capacity beyond 350M (group b) does not yield consistent improvements. The 350M VAE achieves the best ROUGE scores on most datasets, while larger VAEs show marginal gains only on MAUVE. This suggests that REPA is more important than raw VAE capacity for latent space quality.

\paragraph{Latent Channel Dimension.}
Channel dimension 64 (group c) achieves the best results on most metrics, particularly on MAUVE. Lower-dimensional latent spaces appear to benefit the diffusion process by reducing redundancy while retaining sufficient capacity.

\paragraph{REPA Layer Selection.}
Aligning with the 3rd-to-last layer (group d) outperforms aligning with only the last layer. We hypothesize that the final layer's representations are primarily optimized for next-token prediction and may discard information useful for diffusion, whereas intermediate layers retain richer token-level and sentence-level semantics better suited for latent space alignment.

\paragraph{DiT Model Scaling.}
Scaling the DiT from 114M to 768M (group e) yields consistent improvements across all metrics and datasets. We observed that 1M training steps were insufficient for convergence at these scales; due to compute constraints, we extended training to 2M steps for these three configurations to better reveal scaling behavior.

\paragraph{Timestep Schedule.}
The logit-normal schedule with std=1.5 (group f) outperforms both the uniform schedule and logit-normal with std=1.2 on ROUGE metrics across all four datasets, following the SD3~\citep{esser2024scaling} recipe. This confirms that the timestep scheduling insight from visual generation transfers effectively to language modeling.

\paragraph{VAE Reconstruction Accuracy.}

\begin{table}[t]
  \centering
  \begin{minipage}[c]{0.68\columnwidth}
    \centering
    \captionsetup{font=footnotesize}
    \caption{Token reconstruction accuracy (\%) of TextVAE. All VAEs achieve near-perfect reconstruction, indicating that generation quality differences in Table~\ref{tab:ablation} are driven by latent space structure rather than reconstruction fidelity.}
    \label{tab:vae_recon}
    \resizebox{\columnwidth}{!}{
    \begin{tabular}{lcc|cccc}
      \toprule
      VAE Size & ch & REPA & WikiSource & Wikipedia & TinyStories & OBW \\
      \midrule
      \multicolumn{7}{l}{\textit{(a) Latent Channel Dimension}} \\
      350M & 64  & Layer -3 & 97.56 & 97.79 & 99.58 & 99.98 \\
      350M & 128 & Layer -3 & 97.55 & 97.79 & 99.60 & 100.00 \\
      350M & 192 & Layer -3 & 97.55 & 97.78 & 99.60 & 100.00 \\
      \midrule
      \multicolumn{7}{l}{\textit{(b) VAE Model Size}} \\
      350M & 128 & Layer -3 & 97.55 & 97.79 & 99.60 & 100.00 \\
      502M & 128 & Layer -3 & 97.57 & 97.80 & 99.60 & 100.00 \\
      690M & 128 & Layer -3 & 97.57 & 97.80 & 99.60 & 100.00 \\
      \midrule
      \multicolumn{7}{l}{\textit{(c) Effect of REPA}} \\
      350M & 128 & Layer -3 & 97.55 & 97.79 & 99.60 & 100.00 \\
      350M & 128 & --      & 97.56 & 97.78 & 99.60 & 99.99 \\
      \midrule
      \multicolumn{7}{l}{\textit{(d) REPA Layer Selection}} \\
      350M & 128 & Layer -1 & 97.56 & 97.79 & 99.60 & 99.98 \\
      350M & 128 & Layer -3 & 97.55 & 97.79 & 99.60 & 100.00 \\
      \bottomrule
    \end{tabular}
    }
  \end{minipage}%
  \hfill
  \begin{minipage}[c]{0.3\columnwidth}
    \centering
    \captionsetup{font=footnotesize}
    \caption{Sensitivity of CFG scale on TinyStories. Results are reported using 50 denoising steps on 100 TinyStories samples. We employ a 328M DiT and a 350M VAE (chl=128), where the VAE is aligned via REPA to the 3rd-to-last layer of a frozen Qwen3-1.7B. We observe that a CFG scale $w=7$ yields the most favorable generation results.}
    \vspace{10pt}
    \label{tab:cfg}
    \small
    \renewcommand{\arraystretch}{1.1}
    \resizebox{\columnwidth}{!}{
    \begin{tabular}{c|cccc}
      \toprule
      CFG & R-1 & R-2 & R-L & BS \\
      \midrule
      3 & 36.2 & 6.7 & 20.1 & 84.6 \\
      4 & 36.6 & 7.3 & 20.6 & 84.6 \\
      5 & 36.7 & 7.6 & 21.2 & 84.7 \\
      6 & \textbf{37.1} & 7.9 & 21.1 & \textbf{84.9} \\
      \textbf{7} & 37.0 & \textbf{8.1} & \textbf{21.5} & \textbf{84.9} \\
      8 & 36.2 & 7.7 & 21.3 & 84.8 \\
      \bottomrule
    \end{tabular}
    }
  \end{minipage}
\end{table}

Table~\ref{tab:vae_recon} reports the token-level reconstruction accuracy of the TextVAE. All configurations achieve near-perfect accuracy: $\geq$99.6\% on TinyStories and One Billion Words, and $\geq$97.5\% on Wikipedia and WikiSource. The slightly lower accuracy on the latter two is likely due to domain shift: OpenWebText2 primarily consists of Reddit-sourced web content (collected in 2020), whereas Wikipedia and WikiSource are encyclopedic text with different vocabulary distributions, drawn from 2023 snapshots, introducing both topical and temporal distribution gaps. Crucially, the differences across configurations are negligible ($<$0.05\%), yet downstream generation quality varies substantially (Table~\ref{tab:ablation}). This indicates that REPA improves generation not by improving reconstruction, but by shaping the latent space geometry to be more amenable to diffusion modeling.

\paragraph{Classifier-Free Guidance Scale.}

Table~\ref{tab:cfg} shows the effect of classifier-free guidance scale. Performance improves steadily from CFG=3 to CFG=7, peaking around CFG=7 on ROUGE metrics. Higher guidance ($w=8$) causes slight degradation, likely due to reduced diversity.

\section{Limitation}

\methodname{} has several limitations. First, the two-stage training pipeline (VAE then DiT) introduces additional complexity compared to end-to-end AR training. Second, as shown in Table~\ref{tab:vae_recon}, the TextVAE achieves lower reconstruction accuracy on out-of-domain samples (e.g., ${\sim}$97.5\% on Wikipedia and WikiSource), which may propagate errors and limit DiT generation quality on such domains. Expanding the training corpus with more diverse data is expected to mitigate this performance drop.

\section{Conclusion}

We presented \methodname{}, a latent diffusion framework for language modeling that operates entirely in a continuous latent space. By training a TextVAE with Representation Alignment (REPA) and a standard Diffusion Transformer with Flow Matching, \methodname{} achieves state-of-the-art results among diffusion language models while matching autoregressive baselines. A key finding is that the exact recipe proven in visual generation---VAE compression, flow matching, DiT backbone, logit-normal schedule, and classifier-free guidance---transfers effectively to language modeling with minimal architectural modification. This compatibility suggests a path toward unified Diffusion Transformer frameworks that naturally extend across modalities. In future work, we plan to further scale \methodname{} in both data and model size, and to build a unified multimodal architecture that integrates generation and understanding within a single DiT framework.

\bibliographystyle{plainnat}
\bibliography{main}

\newpage
\appendix

This appendix provides supplementary material organized as follows: Appendix~\ref{app:impl_details} describes implementation details including architecture design, training hyperparameters, and compute resources. Appendix~\ref{app:broader_impacts} discusses broader societal impacts. Appendix~\ref{app:future_work} outlines directions for future work. Appendix~\ref{app:step_progression} presents qualitative comparison of different denoising steps.

\section{Implementation Details}
\label{app:impl_details}

\paragraph{TextVAE Architecture.}
The TextVAE encoder and decoder are both standard Transformer models with pre-norm Transformer blocks using LayerNorm and RoPE positional encoding. The encoder takes token embeddings as input and outputs mean and log-variance vectors for each position. The decoder takes sampled latent vectors as input and outputs vocabulary logits for each position.

\paragraph{DiT Architecture.}
The Diffusion Transformer follows the standard DiT architecture~\citep{peebles2023scalable}. No timestep embedding is injected, consistent with LLaDA~\citep{nie2025large} and RADD~\citep{ou2024your}. The clean context latents and noisy target latents are concatenated along the sequence dimension. We use RoPE for positional encoding.

\paragraph{Training Hyperparameters.}
The TextVAE is trained for 200K steps with AdamW optimizer (learning rate 1e-4, weight decay 0.01). The KL weight $\beta$ follows a warmup schedule. The DiTs in ablation study are trained for 1M steps with AdamW (learning rate 1e-4). The DiTs in the main results are trained for 2M steps. For CFG training, we use an unconditional dropout rate of $p_{\text{uncond}} = 0.1$. All experiments are conducted on $8\times$ NVIDIA H200 GPUs with approximately 100K tokens per GPU per mini-batch; VAE training takes approximately 1 day and DiT training takes approximately 2 days.

\section{Broader Impacts}
\label{app:broader_impacts}

This work advances diffusion-based language modeling, a research area with societal implications common to all text generation systems. On the positive side, unifying language and vision generation under a shared diffusion framework could simplify multimodal model development and lower the barrier to building controllable generation systems. However, like all language models, \methodname{} could potentially be used to generate misleading or harmful text. We note that our models are trained on public web text at a modest scale and are not designed for open-ended dialogue or instruction following, which limits the scope of direct misuse. We encourage the community to develop appropriate safeguards as diffusion language models continue to mature.

\section{Future Work}
\label{app:future_work}

Several promising directions remain. \textbf{(1) Scaling laws.}
Investigating how \methodname{} scales to substantially larger model sizes and training corpora---and whether the favorable scaling trends observed in Section~\ref{sec:ablation} continue---is an important next step toward practical diffusion language models. \textbf{(2) Fair evaluation on language understanding benchmarks.} Current benchmarks such as MMLU rely on likelihood-based scoring (e.g., comparing per-token log-probabilities of candidate answers), which inherently favors autoregressive models. Developing evaluation protocols that fairly assess diffusion language models on knowledge and reasoning tasks---for instance, via direct generation and answer extraction---remains an open challenge. \textbf{(3) Surpassing the REPA teacher.} Our TextVAE uses a frozen Qwen3-1.7B as the REPA alignment target. An intriguing question is whether the downstream DiT, by learning to generate in the aligned latent space, can ultimately surpass the representation quality of the teacher model, especially as model and data scale increase. \textbf{(4) Unified understanding and generation.} As discussed in the introduction, \methodname{} demonstrates that the DiT backbone successful in visual generation can be directly applied to language modeling. A natural next step is to build a single DiT-based architecture that handles both vision and language within a shared continuous latent space, unifying generative and understanding capabilities across modalities.

\input{appendix_step_progression}

\end{document}

%% file: appendix_step_progression.tex
\clearpage
\section{Qualitative Examples: Step-by-Step Denoising}
\label{app:step_progression}

We present qualitative examples of \methodname{}'s progressive denoising process on Wikipedia text continuations. Given a conditioning prefix (``Cond''), we show the generated continuation at different diffusion steps (10, 20, 30, 40, 50). These examples illustrate how coherence and factual accuracy emerge progressively as the number of denoising steps increases. 

\begin{table*}[h]
\centering
\caption{Step-by-step denoising progression}
\label{tab:step_progression_2}
\small
\renewcommand{\arraystretch}{1.3}
\begin{tabular}{p{1.5cm}p{12.5cm}}
\toprule
\textbf{Stage} & \textbf{Text} \\
\midrule
\textbf{Cond} & \textit{...In 2014, Gavaris won a Canadian Screen Award for Best Supporting Actor and also won a Constellation Award for best performance in an episode. On August 18, 20 } \\
\midrule
\textbf{Step 10} & 037 she is won Actor Actor Actor Asked Asked Actress following for create Disc\~{a}o Sourceing Jackson starred politics and seem Actress To and Actress and Original andide Actressing Actress.   \\
\midrule
\textbf{Step 20} & 17, he was awarded Canadian Screen Award for Best Actor and Performance for Best drama and lead for BBC America's Best Original Feature film, and a Cored Award for Hiring Actress. \\
\midrule
\textbf{Step 30} & 17, he was awarded Canadian Screen Award for Best Actor and Performance for Best drama and lead for BBC America's Best Original Feature film, and a Cored Award for Hiring Actress. \\
\midrule
\textbf{Step 40} &  17, he was nominated by Queen Turing for Star for Best Supporting Actor and being named one of BBC America and Canada's Space and also won a Constellation Award for best performance. \\
\midrule
\textbf{Step 50} & 17, he was honoured by The Canadian Screen Award for Best Supporting Actor and being named one of BBC America and Canada's Space and also won a Constellation Award for best performance.  \\
\bottomrule
\end{tabular}
\end{table*}

\begin{table*}[h]
\centering
\caption{Step-by-step denoising progression}
\label{tab:step_progression_2}
\small
\renewcommand{\arraystretch}{1.3}
\begin{tabular}{p{1.5cm}p{12.5cm}}
\toprule
\textbf{Stage} & \textbf{Text} \\
\midrule
\textbf{Cond} & \textit{``...also performs on tour with the Julian Siegel Big Band. Recordings Freestone’s debut album with her Trio was In the Chop House, released in 2014 on Whirlwind Recordings. This album featured Freestone on tenor saxophone, Dave Manington on double bass and Tim Giles on drums. The Guardian gave the album 4 stars and said: ``In being supported by only bass’’} \\
\midrule
\textbf{Step 10} & bench state and bass bass more. playing more soundtrack challengingott two and complete diagon agile ballet. composing bass two comment two more state state more more two complete two and develop my dreaming two his soundtrack schedule `` \\
\midrule
\textbf{Step 20} & and bass, the album is to showcase unusual sounds and drums with robust visual improvisational techniques. This trio she featured an imposingly melodic sound and robust composition and lead leader. In the Chop House was her debut for all the albums and hit with excellent albums [\dots] \\
\midrule
\textbf{Step 30} & and bass, the album was to showcase unusual sounds and drums with robust visual improvisational techniques. From behind she showcased an imposingly melodic sound and traditional composition [\dots] In the Chop House was her star for all things on ensemble Jazz: outstanding albums infer Freestone on drums again and it featured Giles on tenor sax with Tim Giles on double bass and Dave Manington on drums. [\dots] \\
\midrule
\textbf{Step 40} & and bass, the album is to embody unusual angles and drums with robust visual improvisational techniques. From behind she created an compensified tenor sound with traditional composition [\dots] In the Chop House was her favorite for all three on ambitious jazz and youth albums. Freestone on drums [\dots] she featured Giles on tenor saxophone Tim Giles on double bass and Dave Manington on drums. [\dots] \\
\midrule
\textbf{Step 50} & and bass, her album is to explore unusual angles and drums with robust visual improvisational techniques. From behind she created an recognisable tenor sound with traditional composition [\dots] In the Chop House was her favorite for all these on ambitious Jazz and youth festivals. Freestone on drums [\dots] featured Giles on tenor saxophone Tim Giles on double bass and Dave Manington on drums. [\dots] \\
\bottomrule
\end{tabular}
\end{table*}

\begin{table*}[h]
\centering
\caption{Step-by-step denoising progression}
\label{tab:step_progression_901}
\small
\renewcommand{\arraystretch}{1.3}
\begin{tabular}{p{1.5cm}p{12.5cm}}
\toprule
\textbf{Stage} & \textbf{Text} \\
\midrule
\textbf{Cond} & \textit{``...the position of Professor of History at the University of Denver, where he remained until his retirement in 1995. At the University of Denver During his time at the University of Denver, Roeder was instrumental in both curriculum development and research program coordination. He served as chair of the History Department during 1985--1986, when the Core Curriculum program was implemented''} \\
\midrule
\textbf{Step 10} & seemIm richser university Pre set here left - 7 more by distinguly - take edition operated recently he me and reasoned3 Experimental `` Henry changing Foundation and - story7 two and and Operating is Oh. during explicitly he taught traditionally programming professor Acad and ranks sailed Technical Professor Modern Professor Scientific studied: found Professor Technical Political Professor in Modern Professor Political Philosophy Judicial Literature taught in History, History of Philosophy [\dots] \\
\midrule
\textbf{Step 20} & . Roeder served a second teaching fellowship later, until 1978. He left Denver, later taking over to the position of Assistant Professor of History at Harvard, where he was instrumental in that process. Roeder was ordained at the University of Chicago, in 1982. His first job was teaching at Harvard University. He returned to history, College of Philosophy, the University of History, and later returned to Harvard University, where he earned his.D. Profites from the North and International Studies. [\dots] He also held a teaching fellowship in history during the Allied II Salvennial of 1933--1986. He then returned to Harvard with history and literature, and a bachelor's thesis on the history of History. [\dots] \\
\midrule
\textbf{Step 30} & . Roeder served a further teaching fellowship until Chicago in 1978. He left that same year in Denver to take the position of Professor in History at Harvard, where he was instrumental in that process. Roeder was elected from the College of William and Mary in1982. His second job was teaching at Harvard University. He returned to history, College of Philosophy, the University of Chicago, and his time at Harvard University, where he earned his.D. in History. [\dots] Roeder held a teaching fellowship during the Counter-Ilennial in 1933--1986. He then returned to Harvard with a teaching fellowship, with a bachelor's thesis at the University of Denver. [\dots] \\
\midrule
\textbf{Step 40} & . Roeder held a second teaching fellowship after Chicago in 1968. He left that same time at Denver to take the position of Professor of History at Harvard, where he was instrumental in that process. Roeder was graduated from the College of William and Mary in1982 with a teaching fellowship in philosophy from Harvard University. He returned to history, College of History, the University of Chicago, and the University of Texas again, where he earned his Dr. of History from the University of Chicago. [\dots] Roeder held a teaching fellowship during the Counter-colonial in 1933\-\-1996. He then returned to Harvard with a teaching fellowship, with a bachelor's thesis at the University of Denver. [\dots] \\
\midrule
\textbf{Step 50} & . Roeder held a second teaching fellowship at Chicago in 1968. He left that same time at Denver to take the position of Professor of History at Harvard, where he was instrumental in curriculum coordination. Roeder was elected from the University of Chicago, in 1982 with a teaching fellowship in history from Harvard University. He returned to history, College of History, the University of Chicago, and the University of Texas, where he then earned his Dr. of History at the University of Denver. [\dots] Roeder held a teaching fellowship during the Counter-colonial from 1933 until 1954. He returned to Harvard with a teaching fellowship, with a bachelor's thesis on the history of Denver. [\dots] \\
\bottomrule
\end{tabular}
\end{table*}

%% file: main.bib
@article{ho2020denoising,
  title={Denoising diffusion probabilistic models},
  author={Ho, Jonathan and Jain, Ajay and Abbeel, Pieter},
  journal={Advances in neural information processing systems},
  volume={33},
  pages={6840--6851},
  year={2020}
}

@article{song2020denoising,
	title={Denoising diffusion implicit models},
	author={Song, Jiaming and Meng, Chenlin and Ermon, Stefano},
	journal={arXiv preprint arXiv:2010.02502},
	year={2020}
}

@article{liu2022flow,
	title={Flow straight and fast: Learning to generate and transfer data with rectified flow},
	author={Liu, Xingchao and Gong, Chengyue and Liu, Qiang},
	journal={arXiv preprint arXiv:2209.03003},
	year={2022}
}

@article{lipman2022flow,
	title={Flow matching for generative modeling},
	author={Lipman, Yaron and Chen, Ricky TQ and Ben-Hamu, Heli and Nickel, Maximilian and Le, Matt},
	journal={arXiv preprint arXiv:2210.02747},
	year={2022}
}

@inproceedings{rombach2022high,
	title={High-resolution image synthesis with latent diffusion models},
	author={Rombach, Robin and Blattmann, Andreas and Lorenz, Dominik and Esser, Patrick and Ommer, Bj{\"o}rn},
	booktitle={Proceedings of the IEEE/CVF conference on computer vision and pattern recognition},
	pages={10684--10695},
	year={2022}
}

@inproceedings{peebles2023scalable,
	title={Scalable diffusion models with transformers},
	author={Peebles, William and Xie, Saining},
	booktitle={Proceedings of the IEEE/CVF international conference on computer vision},
	pages={4195--4205},
	year={2023}
}

@inproceedings{esser2024scaling,
	title={Scaling rectified flow transformers for high-resolution image synthesis},
	author={Esser, Patrick and Kulal, Sumith and Blattmann, Andreas and Entezari, Rahim and M{\"u}ller, Jonas and Saini, Harry and Levi, Yam and Lorenz, Dominik and Sauer, Axel and Boesel, Frederic and others},
	booktitle={Forty-first international conference on machine learning},
	year={2024}
}

@inproceedings{chen2024pixart,
  title={Pixart-$\sigma$: Weak-to-strong training of diffusion transformer for 4k text-to-image generation},
  author={Chen, Junsong and Ge, Chongjian and Xie, Enze and Wu, Yue and Yao, Lewei and Ren, Xiaozhe and Wang, Zhongdao and Luo, Ping and Lu, Huchuan and Li, Zhenguo},
  booktitle={European Conference on Computer Vision},
  pages={74--91},
  year={2024},
  organization={Springer}
}

@article{wan2025wan,
  title={Wan: Open and advanced large-scale video generative models},
  author={Wan, Team and Wang, Ang and Ai, Baole and Wen, Bin and Mao, Chaojie and Xie, Chen-Wei and Chen, Di and Yu, Feiwu and Zhao, Haiming and Yang, Jianxiao and others},
  journal={arXiv preprint arXiv:2503.20314},
  year={2025}
}

@article{li2020optimus,
  title={Optimus: Organizing sentences via pre-trained modeling of a latent space},
  author={Li, Chunyuan and Gao, Xiang and Li, Yuan and Peng, Baolin and Li, Xiujun and Zhang, Yizhe and Gao, Jianfeng},
  journal={arXiv preprint arXiv:2004.04092},
  year={2020}
}

@article{liu2019mu,
  title={$\mu$-forcing: Training variational recurrent autoencoders for text generation},
  author={Liu, Dayiheng and Xue, Yang and He, Feng and Chen, Yuanyuan and Lv, Jiancheng},
  journal={ACM Transactions on Asian and Low-Resource Language Information Processing (TALLIP)},
  volume={19},
  number={1},
  pages={1--17},
  year={2019},
  publisher={ACM New York, NY, USA}
}

@article{dieleman2022continuous,
  title={Continuous diffusion for categorical data},
  author={Dieleman, Sander and Sartran, Laurent and Roshannai, Arman and Savinov, Nikolay and Ganin, Yaroslav and Richemond, Pierre H and Doucet, Arnaud and Strudel, Robin and Dyer, Chris and Durkan, Conor and others},
  journal={arXiv preprint arXiv:2211.15089},
  year={2022}
}

@inproceedings{lin2023text,
  title={Text generation with diffusion language models: A pre-training approach with continuous paragraph denoise},
  author={Lin, Zhenghao and Gong, Yeyun and Shen, Yelong and Wu, Tong and Fan, Zhihao and Lin, Chen and Duan, Nan and Chen, Weizhu},
  booktitle={International Conference on Machine Learning},
  pages={21051--21064},
  year={2023},
  organization={PMLR}
}

@article{wu2023ar,
  title={Ar-diffusion: Auto-regressive diffusion model for text generation},
  author={Wu, Tong and Fan, Zhihao and Liu, Xiao and Zheng, Hai-Tao and Gong, Yeyun and Jiao, Jian and Li, Juntao and Guo, Jian and Duan, Nan and Chen, Weizhu and others},
  journal={Advances in Neural Information Processing Systems},
  volume={36},
  pages={39957--39974},
  year={2023}
}

@article{gulrajani2023likelihood,
  title={Likelihood-based diffusion language models},
  author={Gulrajani, Ishaan and Hashimoto, Tatsunori B},
  journal={Advances in Neural Information Processing Systems},
  volume={36},
  pages={16693--16715},
  year={2023}
}

@inproceedings{han2023ssd,
  title={Ssd-lm: Semi-autoregressive simplex-based diffusion language model for text generation and modular control},
  author={Han, Xiaochuang and Kumar, Sachin and Tsvetkov, Yulia},
  booktitle={Proceedings of the 61st Annual Meeting of the Association for Computational Linguistics (Volume 1: Long Papers)},
  pages={11575--11596},
  year={2023}
}

@article{sahoo2024simple,
  title={Simple and effective masked diffusion language models},
  author={Sahoo, Subham S and Arriola, Marianne and Schiff, Yair and Gokaslan, Aaron and Marroquin, Edgar and Chiu, Justin T and Rush, Alexander and Kuleshov, Volodymyr},
  journal={Advances in Neural Information Processing Systems},
  volume={37},
  pages={130136--130184},
  year={2024}
}

@article{lovelace2023latent,
  title={Latent diffusion for language generation},
  author={Lovelace, Justin and Kishore, Varsha and Wan, Chao and Shekhtman, Eliot and Weinberger, Kilian Q},
  journal={Advances in Neural Information Processing Systems},
  volume={36},
  pages={56998--57025},
  year={2023}
}

@inproceedings{meshchaninovcosmos,
  title={Cosmos: Compressed and Smooth Latent Space for Text Diffusion Modeling},
  author={Meshchaninov, Viacheslav and Chimbulatov, Egor and Shabalin, Alexander and Abramov, Aleksandr and Vetrov, Dmitry},
  booktitle={The Thirty-ninth Annual Conference on Neural Information Processing Systems}
}

@article{shao2025continuous,
  title={Continuous Autoregressive Language Models},
  author={Shao, Chenze and Li, Darren and Meng, Fandong and Zhou, Jie},
  journal={arXiv preprint arXiv:2510.27688},
  year={2025}
}

@article{nie2025large,
  title={Large language diffusion models},
  author={Nie, Shen and Zhu, Fengqi and You, Zebin and Zhang, Xiaolu and Ou, Jingyang and Hu, Jun and Zhou, Jun and Lin, Yankai and Wen, Ji-Rong and Li, Chongxuan},
  journal={arXiv preprint arXiv:2502.09992},
  year={2025}
}

@article{arriola2025block,
  title={Block diffusion: Interpolating between autoregressive and diffusion language models},
  author={Arriola, Marianne and Gokaslan, Aaron and Chiu, Justin T and Yang, Zhihan and Qi, Zhixuan and Han, Jiaqi and Sahoo, Subham Sekhar and Kuleshov, Volodymyr},
  journal={arXiv preprint arXiv:2503.09573},
  year={2025}
}

@article{ou2024your,
  title={Your absorbing discrete diffusion secretly models the conditional distributions of clean data},
  author={Ou, Jingyang and Nie, Shen and Xue, Kaiwen and Zhu, Fengqi and Sun, Jiacheng and Li, Zhenguo and Li, Chongxuan},
  journal={arXiv preprint arXiv:2406.03736},
  year={2024}
}

@article{austin2021structured,
  title={Structured denoising diffusion models in discrete state-spaces},
  author={Austin, Jacob and Johnson, Daniel D and Ho, Jonathan and Tarlow, Daniel and Van Den Berg, Rianne},
  journal={Advances in neural information processing systems},
  volume={34},
  pages={17981--17993},
  year={2021}
}

@article{radford2019language,
  title={Language models are unsupervised multitask learners},
  author={Radford, Alec and Wu, Jeffrey and Child, Rewon and Luan, David and Amodei, Dario and Sutskever, Ilya and others},
  journal={OpenAI blog},
  volume={1},
  number={8},
  pages={9},
  year={2019}
}

@article{yang2025qwen3,
  title={Qwen3 technical report},
  author={Yang, An and Li, Anfeng and Yang, Baosong and Zhang, Beichen and Hui, Binyuan and Zheng, Bo and Yu, Bowen and Gao, Chang and Huang, Chengen and Lv, Chenxu and others},
  journal={arXiv preprint arXiv:2505.09388},
  year={2025}
}

@inproceedings{yu2024representation,
  title={Representation Alignment for Generation: Training Diffusion Transformers Is Easier Than You Think},
  author={Yu, Sihyun and Kwak, Sangkyung and Jang, Huiwon and Jeong, Jongheon and Huang, Jonathan and Shin, Jinwoo and Xie, Saining},
  booktitle={The Thirteenth International Conference on Learning Representations}
}

@article{ho2022classifier,
  title={Classifier-free diffusion guidance},
  author={Ho, Jonathan and Salimans, Tim},
  journal={arXiv preprint arXiv:2207.12598},
  year={2022}
}

@inproceedings{lin2004rouge,
  title={Rouge: A package for automatic evaluation of summaries},
  author={Lin, Chin-Yew},
  booktitle={Text summarization branches out},
  pages={74--81},
  year={2004}
}

@inproceedings{zhang2020bertscore,
  title={BERTScore: Evaluating Text Generation with BERT},
  author={Zhang, Tianyi and Kishore, Varsha and Wu, Felix and Weinberger, Kilian Q and Artzi, Yoav},
  booktitle={International Conference on Learning Representations}
}

@article{pillutla2021mauve,
  title={Mauve: Measuring the gap between neural text and human text using divergence frontiers},
  author={Pillutla, Krishna and Swayamdipta, Swabha and Zellers, Rowan and Thickstun, John and Welleck, Sean and Choi, Yejin and Harchaoui, Zaid},
  journal={Advances in Neural Information Processing Systems},
  volume={34},
  pages={4816--4828},
  year={2021}
}

@misc{Gokaslan2019OpenWeb,  
	title={OpenWebText Corpus},
	author={Aaron Gokaslan and Vanya Cohen},
	howpublished={\url{http://Skylion007.github.io/OpenWebTextCorpus}}, 
	year={2019}
}

@article{pile,
  title={The {P}ile: An 800GB Dataset of Diverse Text for Language Modeling},
  author={Gao, Leo and Biderman, Stella and Black, Sid and Golding, Laurence and Hoppe, Travis and Foster, Charles and Phang, Jason and He, Horace and Thite, Anish and Nabeshima, Noa and Presser, Shawn and Leahy, Connor},
  journal={arXiv preprint arXiv:2101.00027},
  year={2020}
}

@article{eldan2023tinystories,
  title={Tinystories: How small can language models be and still speak coherent english?},
  author={Eldan, Ronen and Li, Yuanzhi},
  journal={arXiv preprint arXiv:2305.07759},
  year={2023}
}

@article{chelba2014one,
  title={One billion word benchmark for measuring progress in statistical language modeling},
  author={Chelba, Ciprian and Mikolov, Tomas and Schuster, Mike and Ge, Qi and Brants, Thorsten and Koehn, Phillipp and Robinson, Tony},
  journal={Interspeech 2014},
  year={2014},
  publisher={ISCA}
}

@article{kingma2013auto,
  title={Auto-encoding variational bayes},
  author={Kingma, Diederik P and Welling, Max},
  journal={arXiv preprint arXiv:1312.6114},
  year={2013}
}

@article{li2022diffusion,
  title={Diffusion-lm improves controllable text generation},
  author={Li, Xiang and Thickstun, John and Gulrajani, Ishaan and Liang, Percy S and Hashimoto, Tatsunori B},
  journal={Advances in neural information processing systems},
  volume={35},
  pages={4328--4343},
  year={2022}
}

@article{team2024chameleon,
  title={Chameleon: Mixed-modal early-fusion foundation models},
  author={Team, Chameleon},
  journal={arXiv preprint arXiv:2405.09818},
  year={2024}
}

@article{wang2024emu3,
  title={Emu3: Next-token prediction is all you need},
  author={Wang, Xinlong and Zhang, Xiaosong and Luo, Zhengxiong and Sun, Quan and Cui, Yufeng and Wang, Jinsheng and Zhang, Fan and Wang, Yueze and Li, Zhen and Yu, Qiying and others},
  journal={arXiv preprint arXiv:2409.18869},
  year={2024}
}
